\documentclass[runningheads]{llncs}
\usepackage{graphicx}
\usepackage{tikz}
\usepackage{comment}
\usepackage{amsmath,amssymb} % define this before the line numbering.
\usepackage{color}
\usepackage[abs]{overpic}
\usepackage{booktabs}
\usepackage{multirow}
\usepackage{diagbox}
\usepackage{array}
\usepackage{enumitem}
\usepackage{csquotes}
\usepackage{floatrow}
\usepackage{scalerel}

\usepackage{wrapfig,lipsum,booktabs}
\usepackage{makecell}

\newcommand{\eg}{e.g.~}
\newcommand{\ie}{i.e.~}
\newcommand{\mytilde}{\raise.17ex\hbox{$\scriptstyle\mathtt{\sim}$}}

\usepackage[accsupp]{axessibility}  % Improves PDF readability for those with disabilities.
\usepackage[pagebackref=true,breaklinks=true,colorlinks,bookmarks=false]{hyperref}

\renewcommand{\vec}[1]{\boldsymbol{#1}}

\newcommand{\set}[1]{\mathcal{#1}}

\newcommand{\jointset}[0]{\set{J}}

\newcommand{\pose}[0]{\vec{\theta}}
\newcommand{\shape}[0]{\vec{\beta}}

\newcommand{\smpl}[0]{M}

\newcommand{\posenet}[0]{f^{\mathrm{enc}}}
\newcommand{\dfnet}[0]{f^{\mathrm{df}}}
\newcommand{\udfnet}[0]{f^{\mathrm{udf}}}

\newcommand{\loss}[0]{\mathcal{L}}

\usepackage[pagebackref=true,breaklinks=true,colorlinks,bookmarks=false]{hyperref}

\DeclareMathOperator*{\argmin}{arg\,min}

\definecolor{onmanifold}{RGB}{56, 118, 29}
\definecolor{offmanifold}{RGB}{116, 27, 71}
\newcommand\lbullet[1][.5]{\mathbin{\ThisStyle{\vcenter{\hbox{%
  \scalebox{#1}{$\SavedStyle\bullet$}}}}}%
}

\begin{document}
\pagestyle{headings}
\mainmatter

\title{\blah: Modeling Human Pose Manifolds with Neural Distance Fields} % Replace with your title

%******************
\newcommand{\blah}{Pose-NDF}
\newcommand{\bvae}{VAE-axis angle}

\titlerunning{\blah}

\author{Garvita Tiwari\inst{1,2} \and
Dimitrije Antić\inst{1} \and
Jan Eric Lenssen\inst{2} \and
Nikolaos Sarafianos\inst{3} \and
Tony Tung\inst{3} \and 
Gerard Pons-Moll\inst{1,2}}

\authorrunning{G. Tiwari et al.}
\institute{University of Tübingen, Germany \\
\email{\{garvita.tiwari, dimirije.antic, gerard.pons-moll\}@uni-tuebingen.de}\\
\and
 Max Planck Institute for Informatics, Saarland Informatics Campus, Germany\\
\email{jlenssen@mpi-inf.mpg.de} \\
\and
Meta Reality Labs Research, Sausalito, USA\\
\email{\{nsarafianos, tony.tung\}@fb.com}}

\maketitle

\begin{abstract}
    We present \blah, a continuous model for plausible human poses based on neural distance fields (NDFs). Pose or motion priors are important for generating realistic new poses and for reconstructing accurate poses from noisy or partial observations. 
%Previous work learns a lower dimensional embedding of human poses using VAEs and use the learned distribution as prior for reconstruction and generating new poses. 
\blah~learns a manifold of plausible poses as the zero level set of a neural implicit function, extending the idea of modeling implicit surfaces in 3D to the high-dimensional domain $SO(3)^K$, where a human pose is defined by a single data point, represented by $K$ quaternions. 
The resulting high-dimensional implicit function can be differentiated with respect to the input poses and thus can be used to project arbitrary poses onto the manifold by using gradient descent on the set of 3-dimensional hyperspheres.
In contrast to previous VAE-based human pose priors, which transform the pose space into a Gaussian distribution, we model the actual pose manifold, preserving the distances between poses.
We demonstrate that \blah{} outperforms existing state-of-the-art methods as a prior in various downstream tasks, ranging from denoising real-world human mocap data, pose recovery from occluded data to 3D pose reconstruction from images. Furthermore, we show that it can be used to generate more diverse poses by random sampling and projection than VAE-based methods. We will
release our code and pre-trained model for further research at \url{https://virtualhumans.mpi-inf.mpg.de/posendf/}.
\end{abstract}

\section{Introduction}
\label{sec:introduction}
\setlength{\belowcaptionskip}{-10pt}
\begin{figure*}[t]
	\centering
	\begin{overpic}[width=0.85\textwidth,unit=1mm]{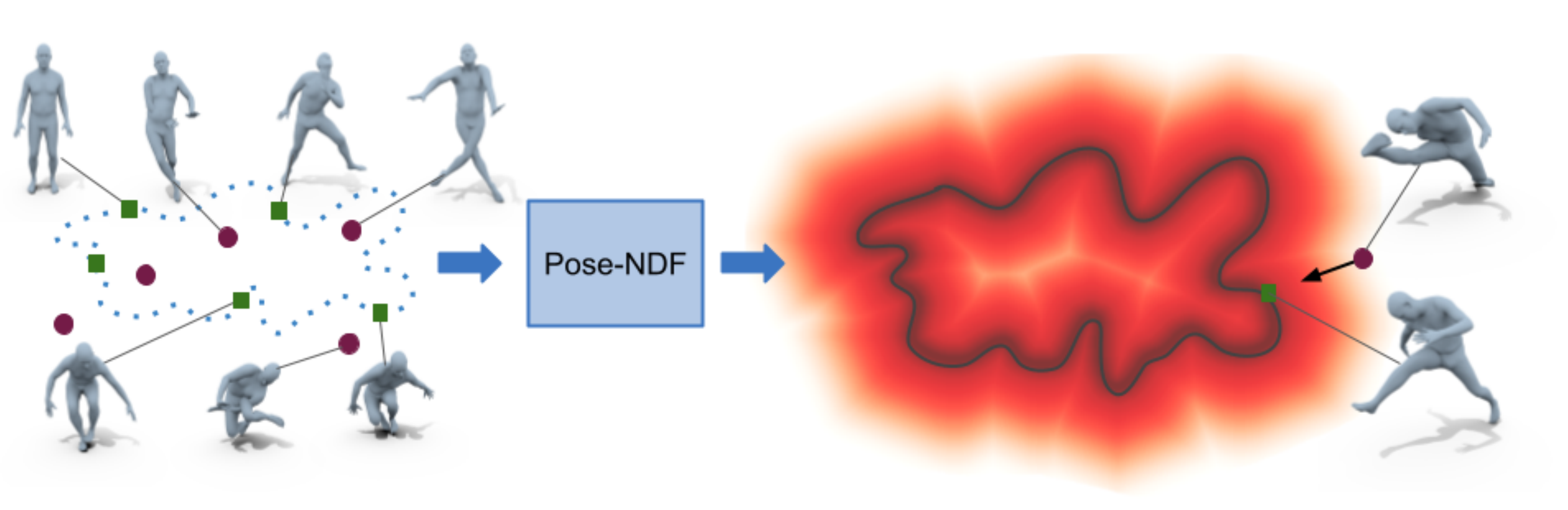}
      \put(30,8){{\parbox{0.2\linewidth}{\scriptsize{$\udfnet = \dfnet\circ\posenet$}}}}
      \put(95,16){{\parbox{0.2\linewidth}{\scriptsize{$\nabla_{\pose} \udfnet$}}}}
    \end{overpic} 
	\caption{We present \blah, a neural unsigned distance field in $SO(3)^K$, which learns the manifold of plausible poses as zero level set. We learn the distance field representation from samples of plausible (\textcolor{onmanifold}{$\blacksquare$}) and unrealistic (\textcolor{offmanifold}{$\lbullet[1.9]$}) poses (\textbf{left}). We encode the input pose (given as a set of quaternions) using a structural MLP $\posenet$ and predict the distance from the joint representation using an MLP $\dfnet$. The gradient $\nabla_{\pose} \udfnet$ and distance value $\udfnet(\pose)$ are used to project implausible poses onto the manifold (\textbf{right}). 
	}
	\label{fig:overview}
\end{figure*}

Realistic and accurate human motion capture and generation is essential for understanding human behavior and human interaction in the scene~\cite{HPS,vip:eccv:2018,PSI:2019,PLACE:3DV:2020,bhatnagar22behave}. Human motion capturing systems, like marker-based systems~\cite{Loper:SIGASIA:2014,KrebsMeixner2021}, IMU-based methods~\cite{HPS,vip:eccv:2018}, or reconstruction from RGB/RGB-D data~\cite{nturgbd2016,chuan2020action2motion,uestc2018,zou2020polarization}, often suffer from artifacts like skating, self-intersections and jitters and produce non-realistic human poses, especially in the presence of noisy data and occlusion. To make the results applicable in fields like 3D scene understanding, human motion generation, or AR/VR applications, it is often required to apply exhaustive manual or automatic cleaning procedures.

In recent years, learned data priors to post-process such non-realistic human poses has become increasingly popular. Prior human pose models mainly focus on learning a joint distribution of individual joints in pose space~\cite{bogo2016keep} or recently in a latent space, using VAEs~\cite{SMPL-X:2019,rempe2021humor,zhang2021learning}. 
They have demonstrated to greatly improve the plausibility of poses after model fitting. However,
VAE-based methods, such as VPoser~\cite{SMPL-X:2019} or HuMoR~\cite{rempe2021humor} make a Gaussian assumption on the space of possible poses, which leads to several limitations: \textbf{1)} They have the \emph{tendency of producing more likely poses} that lie near the mean of the computed Gaussian. Those poses however, might not be the correct ones. \textbf{2)} Distances between individual human poses are \emph{not preserved} in the VAE latent space. Hence, taking small steps towards the Gaussian mean might result in large steps in pose space. \textbf{3)} VAEs have been shown to \emph{fold} a manifold into a Gaussian distribution~\cite{Makhzani:2016}, exposing \emph{dead regions} without any data points in the outer parts of the distribution. Thus, they produce non-plausible samples that are far from the input when traversed in outer regions, as we demonstrate in our experiments. 

To alleviate these issues, we present \blah, a human pose prior that models the full manifold of plausible poses in high-dimensional pose space. We represent the manifold as a surface, where plausible poses lie on the manifold, hence having a zero distance, and non-plausible poses lie outside of it, having a non-zero distance from the surface. We propose to learn this manifold using a high-dimensional neural field, analogously to representing 3D shapes using neural distance fields~\cite{Park_2019_CVPR,chibane2020ndf}. This formulation preserves distances between poses and allows to traverse the pose space along the negative gradient of the distance function, which points to the direction of maximum distance decrease. Using gradient descent in pose space from an initial potentially non-plausible pose, we always find the closest point on the manifold of plausible poses.

An overview of our method is given in Fig.~\ref{fig:overview}. We formulate the problem of learning the pose manifold as a surface completion task in n-dimensional space. In order to learn a pose manifold, there are two key challenges: \textbf{a)} the input space is high-dimensional, and \textbf{b)} the input space is not Euclidean, as it is for 3-dimensional implicit surfaces~\cite{Park_2019_CVPR,chibane2020ndf}.
Instead, the pose space is given as $SO(3)^K$, in which a single pose can be represented by $K$ elements of the rotation group $SO(3)$, describing the orientations of joints in a human body model. To represent group elements, we opted for a quaternion representation, as they are continuous, have an easy-to-compute distance, and are subject to an efficient gradient descent algorithm. We map a given pose to a distance by applying a hierarchical implicit neural function, which encodes the pose based on the kinematic structure of the human body. We train our model using the AMASS dataset~\cite{AMASS:2019}, where each sample from the dataset is treated as a point on the manifold. The learned neural field representation can be used to project any pose onto the manifold, similar to~\cite{chibane2020ndf}. We leverage this property and use \blah{} for diverse pose generation, pose interpolation, as a pose prior for 3D pose estimation from images~\cite{SMPL-X:2019,bogo2016keep}, and motion denoising~\cite{HPS,rempe2021humor}, improving on state-of-the-art methods in all areas.
In summary our contributions are:
\begin{itemize}
	\item [$\bullet$] A novel high-dimensional neural field representation in $SO(3)^K$, \blah{}, which represents the manifold of plausible human poses. 
	\item [$\bullet$] \blah{} improves the state of the art in human body fitting from images by acting as a pose prior. It outperforms other human pose priors, such as VPoser~\cite{SMPL-X:2019} and the human motion prior HuMoR~\cite{rempe2021humor} on motion denoising.
	\item [$\bullet$] Our method is as fast or faster than current state-of-the-art methods, is fully differentiable and the distance from the manifold can be leveraged for finding the optimal step size during optimization.
	\item [$\bullet$] \blah{} generates more diverse samples than previous methods with Gaussian assumptions, which are biased towards generating more likely poses.
\end{itemize}

\section{Related Work}
\label{sec:related_work}
Our method is a \emph{human pose prior} build as \emph{neural field} in high-dimensional space. Thus, we review related work in both of these areas.

\noindent\textbf{Pose and Motion Priors.} 
Human pose and motion priors are crucial for preserving the realism of models estimated from captured data~\cite{vonMarcard2018,HPS,AMASS:2019} and to estimate human pose from images~\cite{bogo2016keep,SMPL-X:2019,xu2019denserac,ExPose:2020,SPIN:ICCV:2019} and videos~\cite{vibe,stoll_videobased}. Further, they can be powerful tools for data generation. Initial work along this direction mainly focused on learning constraints for joint limits in Euler angles~\cite{euler_jointlimit} or swing and twist representations~\cite{range_ballsocket_twist,shao_realisitic,akhter_jointlimits}, to avoid twists and bends beyond certain limits.  
A next iteration of methods fits a Gaussian Mixture Model (GMM) to a pose dataset and uses the GMM-based prior for downstream tasks like image-based 3D pose estimation~\cite{bogo2016keep,sarafianos20163d} or registration of 3D scans~\cite{alldieck19cvpr,bhatnagar2019mgn,tiwari20sizer}. Additionally, simple statistical models, such as PCA, have been proposed~\cite{pca_cyclic,3dpeople_gaussian,Black:ECCV:2000}. With the rise of deep learning and GANs~\cite{Goodfellow2014GenerativeAN}, adversarial training has been used to bring predicted poses close to real poses~\cite{hmrKanazawa17,georgakis2020hierarchical} and for motion prediction~\cite{BarsoumCVPRW2018}. However these are task specific models, HMR~\cite{hmrKanazawa17} models $p(\pose|I)$ and requires an image $I$. HP-GAN~\cite{BarsoumCVPRW2018} models $p(\pose_t|\pose_{t-1})$ and requires pose parameters $\pose_{t-1}$ for previous frame/time. Therefore they cannot be used as a prior for other tasks. 

More recent work uses VAEs to learn pose priors~\cite{SMPL-X:2019}, which can be used for generating pose samples, as prior in pose estimation, or 3d human reconstruction from images or sparse/occluded data. Some works~\cite{rempe2021humor,zhang2021learning,ACTOR:ICCV:2021} propose VAE-based human motion models. HuMoR~\cite{rempe2021humor} proposes to learn a distribution of possible pose transitions in motion sequences using a conditional VAE. ACTOR~\cite{ACTOR:ICCV:2021} learns an action conditioned VAE-Transformer prior. Further work designs pose representations along the hierarchy of human skeletons~\cite{Aksan_2019_ICCV} and uses it for character animation~\cite{andreou2021hierarchyaware}. Concurrent work~\cite{Davydov_2022_CVPR} learns a human pose prior using GANs and highlights the shortcomings of Gaussian assumption based models like VPoser~\cite{SMPL-X:2019}. A VPoser decoder is used as generator (mapping $z \rightarrow \theta$) and an HMR~\cite{hmrKanazawa17}-like discriminator is used to train the model. As described in Sec.~\ref{sec:introduction}, our approach follows a different paradigm than the VAE and GAN-based methods, as we directly model the manifold of plausible poses in high-dimensional space, which leads to a distance-preserving representation.

Before the rise of deep learning, modeling partial pose spaces as implicit functions was common, e.g. as fields on a single shoulder joint quaternion~\cite{Herda02automaticdetermination} or an elbow joint quaternion, conditioned on the shoulder joint~\cite{Herda04hierarchicalimplicit}. However, those ignore the real part of the quaternion, leading to ambiguities in representation, are not differentiable, and are limited to 2 joints in the human body model. In contrast, our method uses a fully differentiable neural network, which learns an implicit surface in higher dimension, taking all human joints and all four components of each quaternion into account.

\noindent\textbf{Neural Fields.} Neural fields~\cite{chibane2020ndf,Park_2019_CVPR,occ_net} for surface modeling have received increasing interest over the recent years. They have been used to model fields in 2D or 3D, representing images or partial differentiable equations~\cite{sitzmann2019siren,igr_amos_icml}, signed or unsigned distances from static 3D shapes~\cite{chibane2020ndf,Park_2019_CVPR,huang2020arch,he2021archrr}, pose-conditioned distance field~\cite{Saito:CVPR:2021,tiwari21neuralgif,LEAP:CVPR:21}, radiance fields~\cite{mildenhall2020nerf,pumarola2020dnerf} and more recently for human-object~\cite{xie22chore,bhatnagar22behave} and hand-object~\cite{zhou2022toch} interactions. For a more detailed overview of neural fields please refer to~\cite{neuralfields}.
Neural fields have recently been brought to higher dimensions to model surfaces in Euclidean spaces~\cite{novello2022neural}. 
In this work, we apply the concept to the high-dimensional, non-Euclidean space of $SO(3)^K$, modeling the unsigned distance to manifolds of plausible human body poses in pose space.

\section{Method}
In this section, we describe our method \blah, a model for manifolds of plausible human poses based on high-dimensional neural fields. We assume that the realistic and plausible human poses lie on a manifold embedded in pose space $SO(3)^K$, with $K$ being the number of joints in the human body. Given a neural network $f: SO(3)^K \mapsto \mathbb{R}^+$, which maps a pose, $\pose \in SO(3)^K$ to a non-negative scalar, we represent the manifold of plausible poses as the zero level set:
\begin{equation}
\mathcal{S} = \{\pose \in SO(3)^K \mid f(\pose) = 0\} \textrm{,}
\end{equation}
such that the value of $f$ represents the unsigned distance to the manifold, similar to neural fields-based 3D shape learning~\cite{Chabra2020,chibane2020ndf,igr_amos_icml,Park_2019_CVPR}.
Without loss of generality, we use the SMPL body model~\cite{SMPL:2015,SMPL-X:2019}, resulting in poses $\pose$ with $K =21$ joints. 

\subsection{Quaternions as Representation of SO(3)}
\label{sec:quaternions}
A human pose is represented by 3D rotations of individual joints in the human skeleton. The $3$-dimensional rotation group $SO(3)$ has several common vector space representations that are used to describe group elements in practice. Frequently used examples are rotation matrices, axis-angle representations or unit quaternions~\cite{Huynh:2009}. \blah~requires the representation to have specific properties: a) we aim to model a \emph{manifold}, continuously embedded in pose space. Thus, the chosen representation should be continuous in parameter space, which prohibits axis-angle representations; b) the representation should enable efficient computation of the geodesic distance between two elements; c) our algorithm requires \emph{efficient gradient descent} in pose space. As described in Sec.~\ref{sec:projection}, quaternions are subject to such a gradient descent algorithm that makes use of the efficient reprojection to $SO(3)$ by vector normalization. In contrast, rotation matrices would require more expensive orthogonalization. Therefore, we chose unit quaternions as the best-suited $SO(3)$ representation of joints, as they fulfill all three properties. We will use $\pose =\{\pose_1,...,\pose_K\}$ to denote the quaternions for all $K$ joints of a pose. 
Each quaternion represents the rotation of a joint with respect to its parent node. Since quaternions lie on $S^3$ (embedded in $4$-dimensional space) the full pose $\pose$ can be easily used as input for a neural network $f: \mathbb{R}^{4 K} \rightarrow \mathbb{R}^+$. We define the distance $d:(S^3)^K \times (S^3)^K \rightarrow \mathbb{R}^+$ between two poses $\pose = \{\pose_1,...,\pose_K\}$ and $\hat{\pose} = \{\hat{\pose}_1, ..., \hat{\pose}_K\}$ as:
\begin{equation}
    d(\pose, \hat{\pose}) = \sqrt{\sum_{i=1}^K \frac{w_i}{2} (\arccos{|\pose_i^\top\cdot \hat{\pose}_i|})^2} \textrm{,}
    \label{eq:quatdistance}
\end{equation}
where the individual elements of summation are a metric on $SO(3)$~\cite{Huynh:2009} and $w_i$ is the weight associated with each joint based on their position in the kinematic structure of the SMPL body model (\ie early joints in the chain have higher weights). It should be noted that the double cover property of unit quaternions, that is, the quaternions $\mathbf{q}$ and $-\mathbf{q}$ represent the same $SO(3)$ element, does not lead to additional challenges. We simply train the network to be point symmetric by applying sign flip augmentation on input quaternions.

\subsection{Hierarchical Implicit Neural Function} \label{sec:poseenc}
We represent the human pose with quaternions in local coordinate frames of the parent joint, using the kinematic structure of the SMPL body model. We treat the joints in local coordinate frame, so that continuous manipulation of a single joint corresponds to realistic motion. However, this might result in unrealistic combination of rotation of joints. The plausibility of individual joints depends on the ancestor rotations and thus needs to be conditioned on them. In order to incorporate this dependency, we use a hierarchical network $\posenet$, which encodes the human pose based on the model structure~\cite{Aksan_2019_ICCV,georgakis2020hierarchical,LEAP:CVPR:21}, before predicting the distance based on the joint representation. 

Formally, for a given pose $\pose = \{\pose_1,...,\pose_K\}$, where $\pose_k$ is the pose for joint $k$, and a function $\tau(k)$, mapping the index of each joint to its parent joints index, we encode each pose using an MLP as:
\begin{equation}
   \posenet_1 :(\pose_1) \mapsto \mathbf{v}_1 \; \; \; \; \;  \posenet_k :(\pose_k,\mathbf{v}_{\tau(k)}) \mapsto \mathbf{v}_k \textrm{,} \quad k \in \{2 \hdots K\}
\end{equation}
which takes the quaternion pose and encoded feature $\mathbf{v}_{\tau(k)} \in \mathbb{R}^l$ of its parent joint as input and generates $\mathbf{v}_k \in \mathbb{R}^l$, where $l$ is the dimension of feature. We then concatenate the encoded feature for every joint to get a combined pose embedding $\mathbf{p} = [\mathbf{v}_1 || \hdots || \mathbf{v}_K]$. This embedding is processed by an MLP $\dfnet:\mathbb{R}^{l\cdot K} \rightarrow \mathbb{R}^+$, which predicts the unsigned distance for the given pose representation $\mathbf{p}$. Collectively the complete model $\udfnet(\pose) = (\dfnet \circ\posenet)(\pose)$, is termed as \blah{}, where  $\udfnet: SO(3)^K \mapsto \mathbb{R}^+$.

\subsection{Loss functions}
\label{sec:training}
We train the hierarchically structured neural field $\udfnet$ to predict the geodesic distance to the plausible pose manifold for a given pose. The training data is given as a set $\mathcal{D} = \{(\pose_i, d_i)\}_{1\leq i \leq N}$, containing pairs of poses $\pose_i$ and distances $d_i$ (Eq.~\ref{eq:quatdistance}). We train the network with the standard distance loss $\loss_{\mathrm{UDF}}$, and an Eikonal regularizer $\loss_{\mathrm{eikonal}}$, which encourages a unit-norm gradient for the distance field outside of the manifold~\cite{eikonal_org,igr_amos_icml}: 
\begin{equation}
\label{eq:loss}
 \loss_{\mathrm{UDF}} = \sum_{(\pose,d)\in \mathcal{D}}||\udfnet(\pose) - d_{\pose}||_{2} \;  \; \; \loss_{\mathrm{eikonal}} = \sum_{\substack{(\pose,d)\in \mathcal{D},\;d\neq 0}}(|| \nabla_{\pose}\udfnet(\pose)|| -1 )^2 \textrm{,} 
\end{equation}

More details about training data, network architecture is provided in the supplementary material.

\subsection{Projection Algorithm}
\label{sec:projection}
Given a trained model $\udfnet$, it can be applied to project an arbitrary pose $\pose$ to the manifold of plausible poses. We use the predicted distance $\udfnet(\pose)$ and gradient information $\nabla_{\pose}\udfnet(\pose)$ to project a query pose to the manifold surface $ \mathcal{S}$, as was previously done in unsigned distances functions for 3D shapes~\cite{chibane2020ndf}. In our case, given $SO(3)$ poses, this amounts to finding:
\begin{equation}
\label{eq:data}
\hat{\pose} = \argmin_{\pose \in SO(3)^K} d(\pose, \mathcal{S}) \textrm{,}
\end{equation}
where $d(\pose, \mathcal{S})$ is the distance (Eq.~\ref{eq:quatdistance}) of $\pose$ to the closest point in $\mathcal{S}$.
We find $\hat{\pose}$ by applying gradient descent on the $3$-sphere, using gradient information $\nabla_{\pose}f(\pose)$ and distances $f(\pose)$, obtained from the implicit neural function $f$. One step is given as:
\begin{equation}
\label{eq:proj}
    \pose^i = \pose^{i-1} - \alpha f(\pose^{i-1})\nabla_{\pose}f(\pose^{i-1}) \textrm{,}
\end{equation}
followed by a re-projection to the sphere (\ie vector normalization) after several iterations. This algorithm is guaranteed to converge to local minima on the sphere, which in our case, assuming a correctly learned distance function, is the nearest point on the pose manifold.

\section{Experiments and Results}
 In this section we evaluate \blah{} and show the different use cases of our pose model, which include the ability to serve as a \emph{prior in denoising motion sequences or recovery} from partial observations (Sec.~\ref{sec:denoise}), \emph{prior for recovering plausible poses} from images (Sec.~\ref{sec:3dpose}) using an optimization-based method, \emph{pose generation} (Sec.~\ref{sec:pose_gen}) and \emph{pose interpolation} (Sec.~\ref{sec:pose_inter}). We demonstrate that the \blah{} method \emph{outperforms the state-of-the-art VAE-based human pose prior methods}. We also show the \emph{advantages} of our distance field formulation over VAEs or Gaussian assumption models (Sec.~\ref{sec:vae_gan}). Before turning to the results, we explain training and implementation details of {\blah} in Sec.~\ref{sec:experimentalsetup}. 

\subsection{Experimental Setup}
\label{sec:experimentalsetup}

We use the AMASS dataset~\cite{AMASS:2019} for training. As mentioned in Sec.~\ref{sec:training}, we train the network with supervision on predicted distance values, and hence we create a dataset of pose and distance pairs ($\pose, d_{\pose}$). Since the training samples from AMASS lie on the desired manifold, $d_{\pose} = 0$ is assigned to all poses in the dataset. We then randomly generate negative samples with distance $d_{\pose} > 0$ by adding noise to AMASS poses. Please find details of data preparation in supplementary. We train our model in a \textbf{multi-stage} regime by varying the type of training samples used. We start our training with manifold poses $\pose_m$ and non-manifold poses $\pose_{nm}$ with a large distance to the desired manifold. Then we increase the number of non-manifold poses $\pose_{nm}$ with a small distance in each training batch. This training scheme helps to initially learn a smooth surface and to iteratively introduce the fine details over the course of training.
Our \textbf{network architecture} consists of one 2-layer MLP $\posenet$ with an output feature size of $l=6$ for each joint, similar to~\cite{LEAP:CVPR:21}. Thus, the pose encoding network generates a feature vector $\mathbf{p} \in \mathbb{R}^{126}$. We implement the distance field network $\dfnet$ as a 5-layer MLP. For training, we use the softplus activation in the hidden layer and train the network end-to-end using the loss functions described in Eq.~\eqref{eq:loss}. 

\begin{table}[t]
\setlength{\tabcolsep}{1.0em}
\centering
\caption{\textbf{Motion denoising}: We compare the per-vertex error (in $cm$) on mocap data from HPS (\textbf{left}) and AMASS (\textbf{middle}) and on artificially created noisy AMASS data (\textbf{right}). In all cases, \blah{} based motion denoising results in the least error. We also observe that in case of mocap data (HPS, AMASS), motion denoising using \blah{} results in very small error (small change from input), which is the desired behavior as these mocap poses are already realistic and hence close to our learned manifold. On the other hand, HuMoR changes the input pose significantly.
} 
\resizebox{\textwidth}{!}{
\begin{tabular}{lccccccccc}
\toprule
Data  & \multicolumn{3}{c}{HPS~\cite{HPS}} & \multicolumn{3}{c}{AMASS~\cite{AMASS:2019}} & \multicolumn{3}{c}{Noisy AMASS}\\
\cmidrule(r){2-4}\cmidrule(r){5-7} \cmidrule(r){8-10}
\multicolumn{1}{l}{\# frames} &  60   & 120  &  240 &  60  & 120  &  240 &  60   & 120  &  240    \\
\midrule
Method &  &  & \\
\midrule
VPoser~\cite{SMPL-X:2019}    &  4.91 & 4.16 & 3.81  &  1.52 & 1.55    & 1.47     &  8.96  & 9.13  & 9.15   \\
HuMoR~\cite{rempe2021humor}  &  9.69 & 8.73 & 10.86  &  3.21 & 3.62 & 3.67  &  11.04 & 17.14 & 30.31  \\
\textbf{\blah{}}                          &  \textbf{2.32} & \textbf{2.14} & \textbf{2.11}  &  \textbf{0.59} & \textbf{0.55} & \textbf{0.54}  &  \textbf{7.96} & \textbf{8.31} & \textbf{8.46}   \\
\bottomrule
\end{tabular}
\label{tab:motion_denoise}
}
\end{table}

\subsection{Denoising Mocap Data}
\label{sec:denoise}
Human motion capture has been done using diverse setups ranging from RGB, RGB-D to IMU based capture systems. The data captured from these sources often produce artifacts like jitters, unnaturally rigid joints or weird bends at some joints, or positions with only partial observations. Prior work~\cite{rempe2021humor} improves the quality of captured motion sequences by using an optimization-based method, with the goal of recovering the captured data and preserving the realism of human poses. A robust and expressive human pose prior is key to preserve the realism of optimized poses, along with preserving the original data. Following HuMoR~\cite{rempe2021humor}, we demonstrate the effectiveness of our pose manifold for: 1) motion denoising and 2) fitting to partial data.

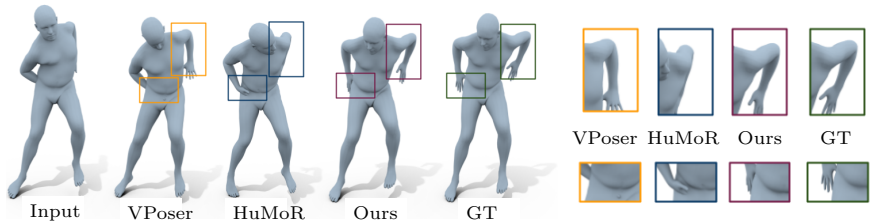
\begin{figure}[t]
	\centering
	\begin{tabular}{ c}
				\begin{overpic}[width=0.6\textwidth,unit=1mm]{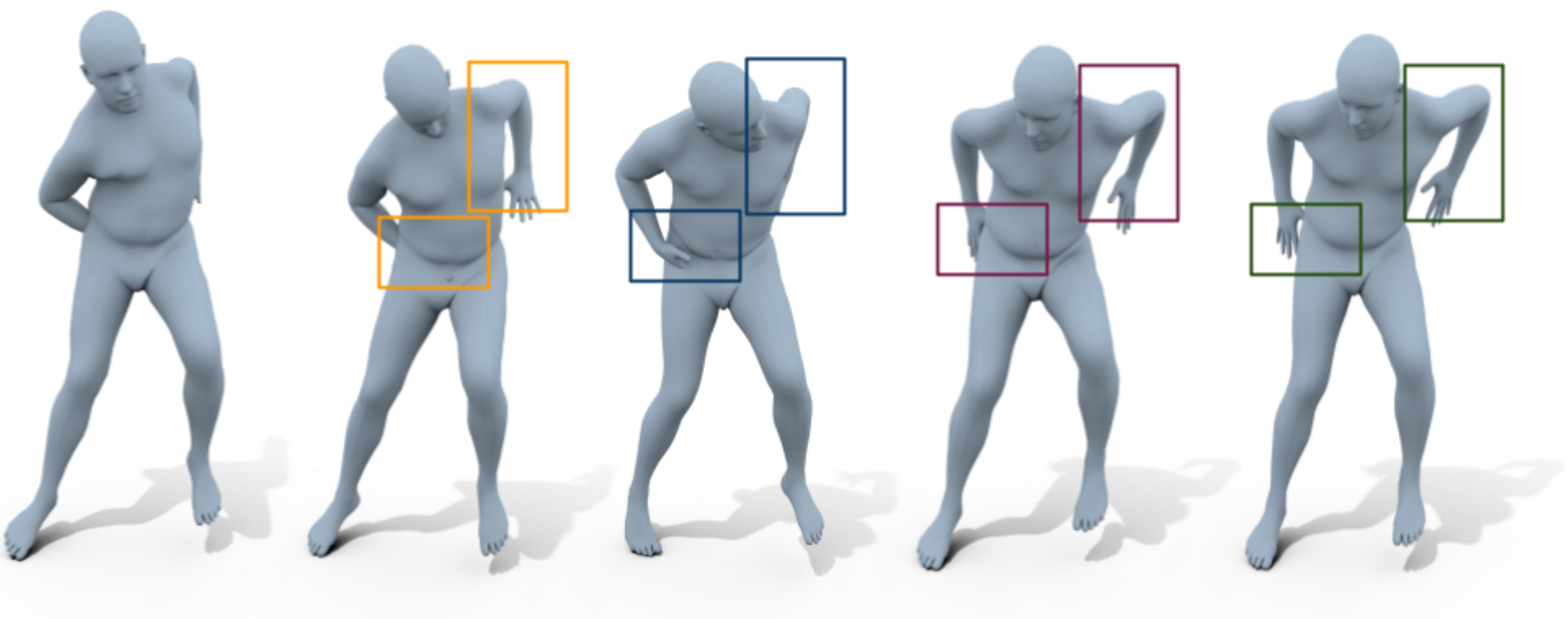}
     		\put(2,0){\colorbox{white}{\parbox{0.05\linewidth}{%
     \scriptsize{Input}}}}
          		\put(15,0){\colorbox{white}{\parbox{0.05\linewidth}{%
     \scriptsize{VPoser}}}}
  	    \put(29,0){\colorbox{white}{\parbox{0.05\linewidth}{%
     \scriptsize{HuMoR}}}}
       	    \put(45,0){\colorbox{white}{\parbox{0.05\linewidth}{%
     \scriptsize{Ours}}}}
        	    \put(60,0){\colorbox{white}{\parbox{0.05\linewidth}{%
     \scriptsize{GT}}}}
\end{overpic}

	\begin{overpic}[width=0.35\textwidth,unit=1mm]{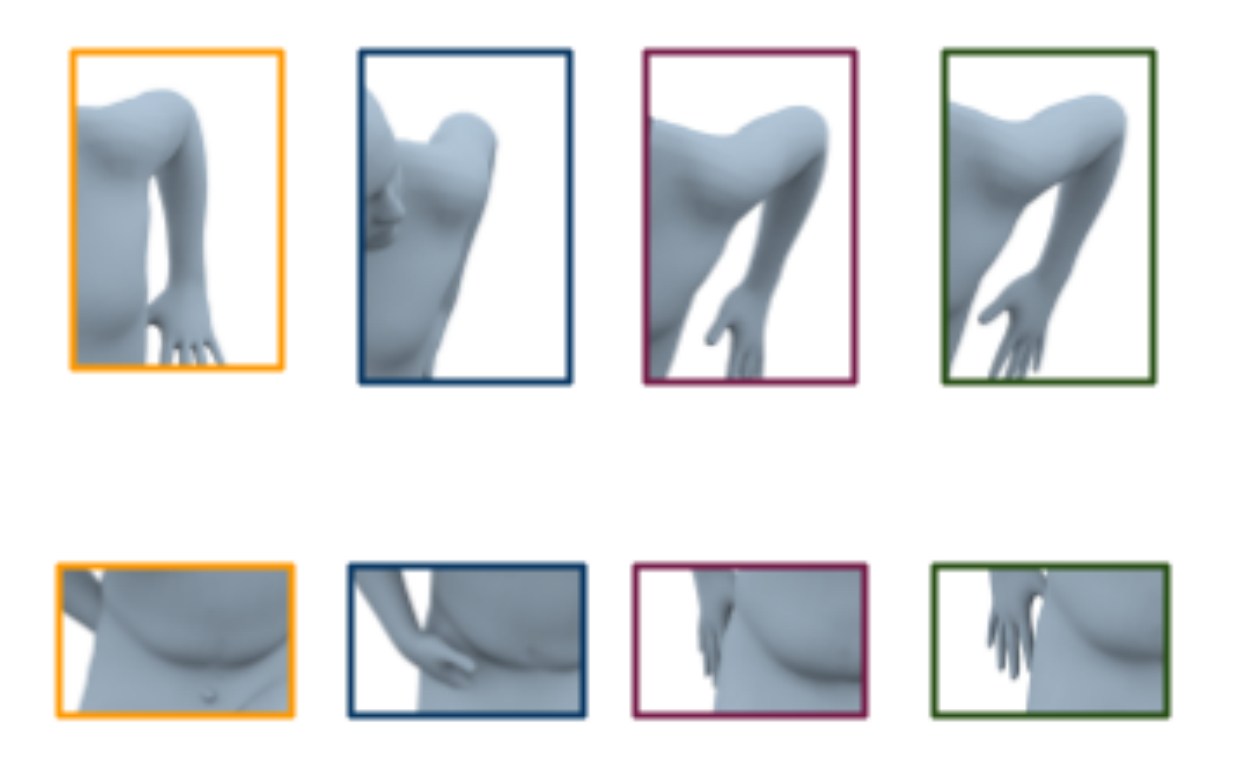}
     		\put(0,10){\colorbox{white}{\parbox{0.05\linewidth}{%
     \scriptsize{VPoser}}}}
          		\put(10,10){\colorbox{white}{\parbox{0.05\linewidth}{%
     \scriptsize{HuMoR}}}}
       	    \put(22,10){\colorbox{white}{\parbox{0.05\linewidth}{%
     \scriptsize{Ours}}}}
        	    \put(33,10){\colorbox{white}{\parbox{0.05\linewidth}{%
     \scriptsize{GT}}}}
 
\end{overpic}

\end{tabular}
	\caption{\textbf{Motion denoising}: We observe that \blah{} based motion denoising makes the pose realistic and  solves small intersection issues, while VPoser and HuMoR still result in unrealistic poses. }
	\label{fig:denoise_amass}
\end{figure}

We follow the same experimental setup as~\cite{rempe2021humor}, but only deal with human poses and thus, remove the terms corresponding to human-scene contact and translation of root joint. In total, we find the pose parameters $\hat{\theta}^t$ at frame t as:
\begin{equation}
\label{eq:denoise}
\hat{\pose}^t = \argmin_{\pose} \lambda_{\mathrm{v}} \mathcal{L}_\mathrm{v} + \lambda_{\pose}\mathcal{L}_{\pose} + \lambda_{t}\mathcal{L}_{t} \textrm{,}
\end{equation}
where $\mathcal{L}_\mathrm{v}$ makes sure that the optimized pose is close to the observation and the temporal smoothness term $\mathcal{L}_{t}$ enforces temporal consistency:
\begin{equation}
\label{eq:denoise_ind}
\mathcal{L}_\mathrm{v} = ||\jointset(\shape_{0},\hat{\pose}^t) -  \jointset_\mathrm{obs} ||_2^2   \;  \;  \; \;  \; \mathcal{L}_{t}= ||\smpl (\shape_{0},\hat{\pose}^\mathrm{t}) -  \smpl (\shape_{0},\pose^\mathrm{t-1}) ||_2^2 \textrm{.}
\end{equation}
Here, $\jointset(\shape,\pose)$ represent vertices (mocap markers) and $\smpl(\shape,\pose)$ represents SMPL mesh vertices for a given pose ($\pose$) and shape ($\shape$) parameters of SMPL~\cite{SMPL:2015}.
Finally, we use \blah{} as a pose prior term in the optimization by minimizing the distance of the current pose from our learned manifold, $\mathcal{L}_{\pose} = \udfnet(\pose)$. We leverage the distance $\udfnet(\pose)$ to get the optimal step size during optimization.

\begin{figure}[t]
	\centering

	\begin{overpic}[width=0.88\textwidth,unit=1mm]{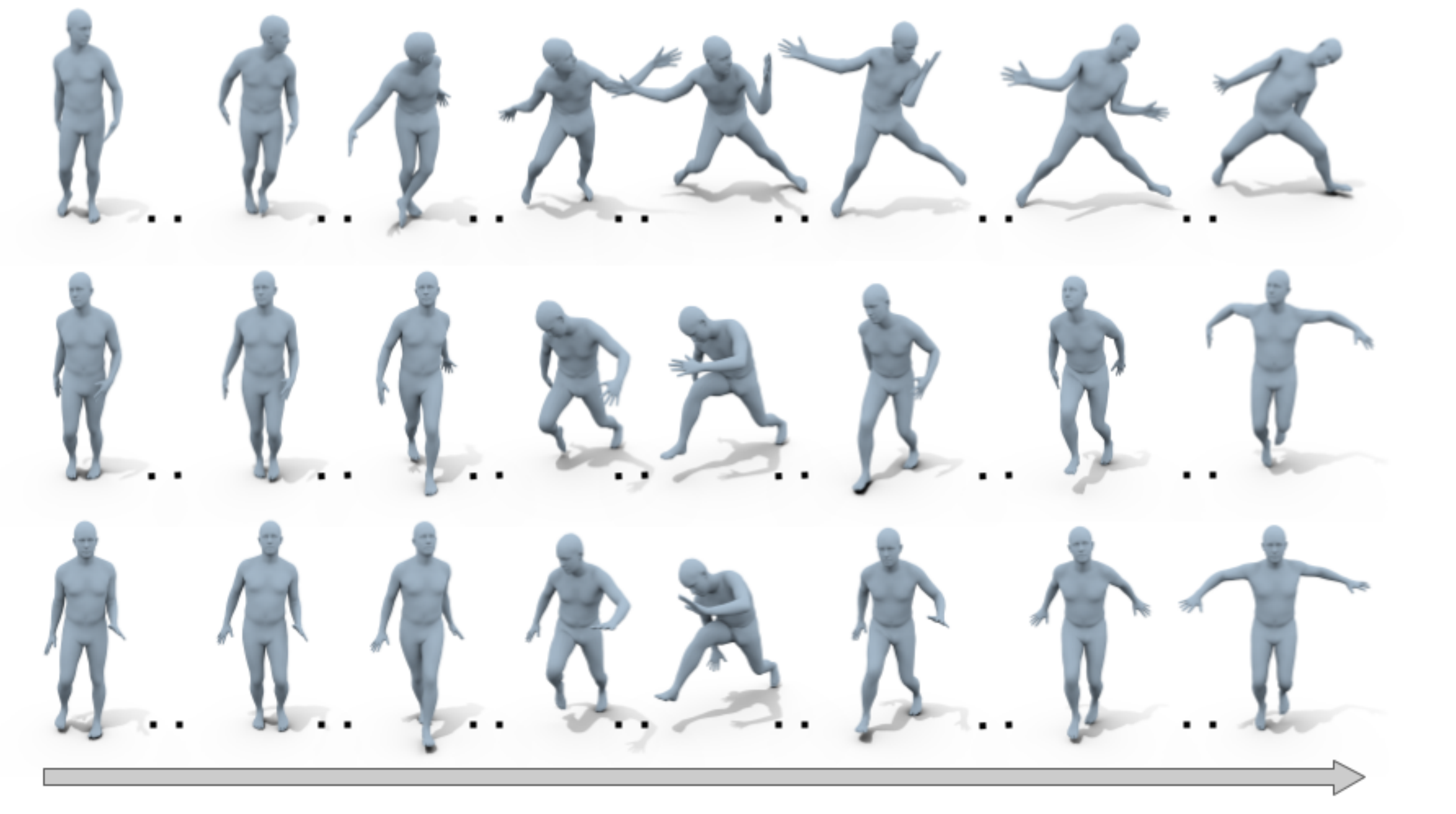}
     		\put(-7,50){\colorbox{white}{\parbox{0.05\linewidth}{%
     \scriptsize{HuMoR}}}}
          		\put(-7,30){\colorbox{white}{\parbox{0.05\linewidth}{%
     \scriptsize{GT}}}}
       	    \put(-7,10){\colorbox{white}{\parbox{0.05\linewidth}{%
     \scriptsize{Ours}}}}
       	    \put(10,-0.7){\colorbox{white}{\parbox{0.05\linewidth}{%
     \scriptsize{Time}}}}
    \end{overpic}
    \caption{\textbf{Motion denoising}: We compare the results on motion denoising using \blah{} and HuMoR~\cite{rempe2021humor} as priors with GT data and visualize every \(10^{th}\) frame of a sequence. We observe that for HuMoR (\textbf{top}) the correction in input pose accumulates over time and makes the output pose significantly different from the GT (\textbf{middle}). \blah{} remains close to observations while correcting unrealistic poses (\textbf{bottom}). }
	\label{fig:denoise_humor}
\end{figure}
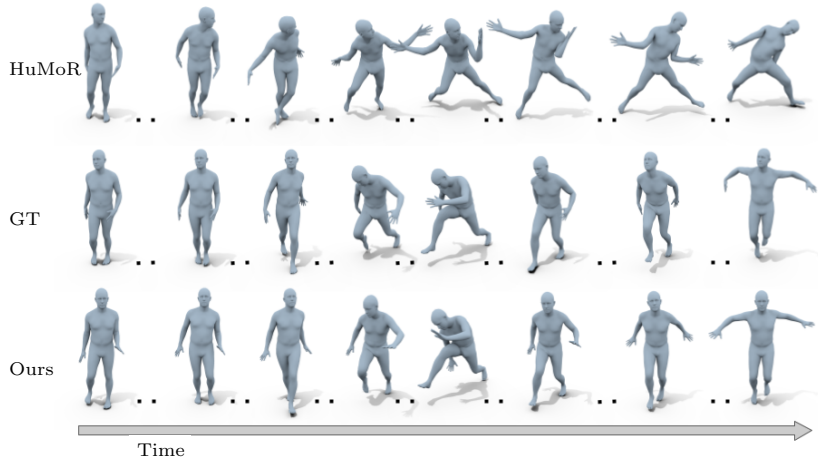

We evaluate on two different settings: 1) clean mocap datasets and 2) a noisy mocap dataset. For clean mocap datasets, we use HPS~\cite{HPS} and the test split of AMASS~\cite{AMASS:2019,SMPL-X:2019}. For the noisy mocap dataset, we create random noisy sequences by adding Gaussian noise to AMASS test sequences and call it “Noisy AMASS”. The average noise introduced in “Noisy AMASS” is 9.3 $cm$.
We use a list of SMPL mesh vertices $\jointset$ as observation during optimization. We created the data with a fixed shape and do not optimize for shape parameters $\shape$. Instead of adding noise to joint locations, we add noise directly to the rotation of each joint. This is done for all methods to ensure a fair comparison. For HuMoR~\cite{rempe2021humor}, we use the \emph{TestOpt} optimization from the original work. VPoser does not have motion experiments, which is why we combine the latent space optimization from the original work with our optimization given in Eq.~\eqref{eq:denoise} to ensure that we compare against the best possible result. Specifically, we first encode the rotation matrix representation of noisy input pose $\pose^t$ using the VPoser encoder as $\vec{z}^t = f_\mathrm{v\_enc}(\pose^t)$, then add random noise ($\hat{\epsilon}$) in the latent space and reconstruct the pose by $\tilde{\pose}^\mathrm{t} = f_\mathrm{v\_dec}(\vec{z}^t + \hat{\epsilon})$. Following~\cite{zhang2021learning}, we observe that the temporal term in latent space yields better results than the temporal term in input pose/vertices, which we used in the VPoser experiment. The prior and temporal term for VPoser-based denoising are given as:
\begin{equation}
\label{eq:vposer_denoise}
 \mathcal{L}_{\pose}^\textrm{VPoser} = || \hat{\epsilon}||_2  \;  \; \;  \;  \;  \; \mathcal{L}_{t}^\textrm{VPoser} = || \vec{z}^{t-1} - \hat{\vec{z}}^t||_2 .
\end{equation}

\begin{table}[t]
\setlength{\tabcolsep}{1.0em}
\centering
\caption{\textbf{Motion estimation from partial 3D observations}: We compare per-vertex error (in $cm$). It can be seen that for leg and arm/hand occlusions, \blah{} reconstructs the pose better than VPoser and HuMoR. For occluded shoulders, HuMoR takes the lead. We observe that results of \blah{} depend on the initialization of the occluded joint, as it is expected from manifold projection. } 
\resizebox{\textwidth}{!}{
\begin{tabular}{lccccccccc}
\toprule
Data & \multicolumn{3}{c}{Occ. Leg} & \multicolumn{3}{c}{Occ. Arm+hand} & \multicolumn{3}{c}{Occ. Shoulder +Upper Arm}\\
\cmidrule(r){2-4}\cmidrule(r){5-7} \cmidrule(r){8-10}
\multicolumn{1}{c}{\# frames} &  60   & 120  &  240 &  60  & 120  &  240 &  60   & 120  &  240    \\
\midrule
Method &  &  & \\
\midrule
VPoser~\cite{SMPL-X:2019}  &  2.53 & 2.57 & 2.54  &  8.51 & 8.52 & 8.59  &  9.98 & 9.49 & 9.48  \\
HuMoR~\cite{rempe2021humor}   &  5.60 & 6.19 &  9.09  &  7.83 & 8.44 & 10.25  &  \textbf{4.75} & \textbf{5.11} & \textbf{4.95}  \\
\textbf{\blah{}}         &  \textbf{2.49} & \textbf{2.51} & \textbf{2.47}  &  \textbf{7.81} & \textbf{8.13} &  \textbf{7.98}  &  7.63 & 7.89 & 6.76   \\
\bottomrule
\end{tabular}
\label{tab:motion_partial}
}
\end{table}

\noindent
\textbf{Results.} We compare motion denoising between HuMoR~\cite{rempe2021humor} (\emph{TestOpt}), Eq.~\eqref{eq:denoise} with VPoser prior~\cite{SMPL-X:2019}, and Eq.~\eqref{eq:denoise} with \blah{} prior in Table~\ref{tab:motion_denoise}.
% and we will use the keywords HuMoR~\cite{rempe2021humor}, VPoser and \blah{}, respectively. 
\blah{} achieves the lowest error in all settings. 
For mocap datasets like AMASS and HPS the motion is realistic, but can have small artifacts and jitter. Thus, an ideal motion/pose prior should not change the overall pose of these examples, but only fix these local artifacts. 
We observe that, numerically, VPoser and \blah{}-based optimization do not change the input pose significantly. However HuMoR changes the pose and this change increases with an increasing number of frames. This is because HuMoR is a motion-based prior (conditioned on the previous pose) and, hence, over time the correction in pose accumulates and makes the output pose significantly different from the input.

For the ``Noisy AMASS" data, \blah{}-based optimization outperforms prior work. We visualise the denoising results in Fig.~\ref{fig:denoise_amass}, and observe that the \blah{}-based method produces realistic and close to GT results. We further compare results of a sequence with HuMoR in Fig.~\ref{fig:denoise_humor}. HuMoR results in large deviations from the input/GT, due to accumulation of correction over time.

\noindent \textbf{Fitting to partial data.} We use the test set of AMASS and randomly create occluded poses (\eg missing arm or legs or shoulder joint) and quantitatively compare with HuMoR~\cite{rempe2021humor} and VPoser~\cite{SMPL-X:2019} in Table~\ref{tab:motion_partial}. 
We use Eq.~\eqref{eq:denoise} for VPoser and \blah{}-based optimization. We only optimize for the occluded joints and for our model, we initialize the occluded joint pose randomly (close to 0). For HuMoR, we use the \emph{TestOpt} provided in their paper. 
We evaluate on three different type of occlusions: 1) occluded left leg, 2) occluded left arm and 3) occluded right shoulder and upper arm. For the occluded leg case, VPoser and our prior-based method perform better. We believe this is because the majority of the poses in both AMASS training and test are upright with nearly straight legs and hence VPoser is biased towards these poses. For our method, it highly depends on initialization. Since we have used an initialization close to rest position, our optimization method generates smaller error for occluded legs but higher errors for occluded arms and shoulders, as they usually are more far away from the rest pose. 
For HuMoR, the motion generated is realistic and plausible, but in some cases results in large deviation from ground truth, because the correction in input pose accumulates over the time.

\begin{figure}
\begin{overpic}[width=0.48\textwidth,unit=1mm]{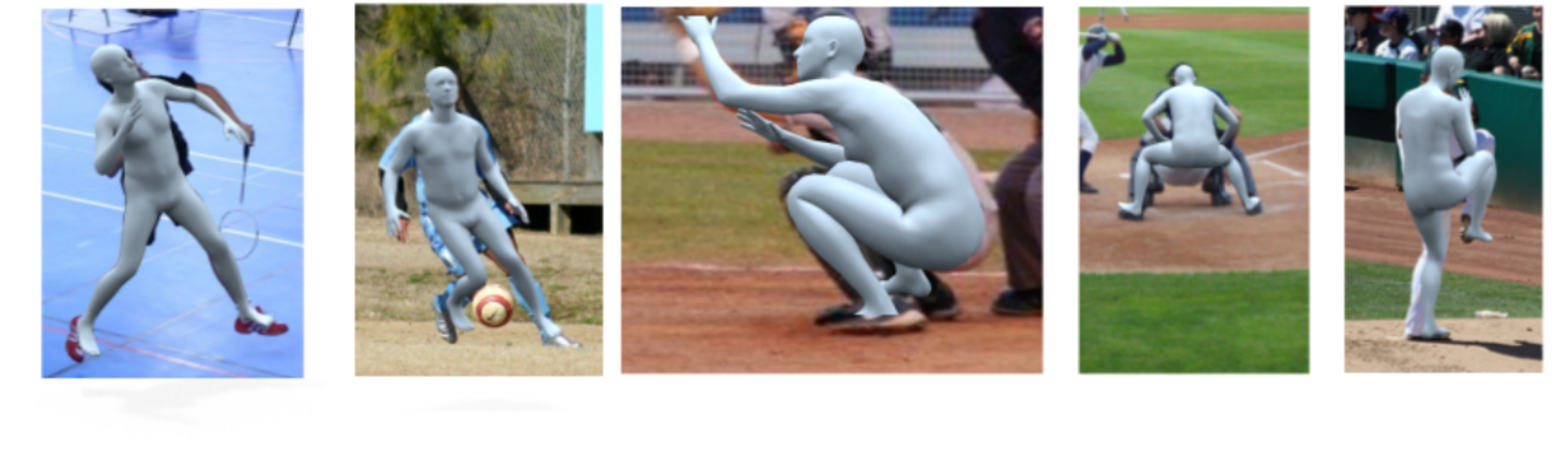}
	\put(15,1){{\parbox{0.5\linewidth}{%
     LSP dataset~\cite{Johnson10}}}}

\end{overpic}
\begin{overpic}[width=0.48\textwidth,unit=1mm]{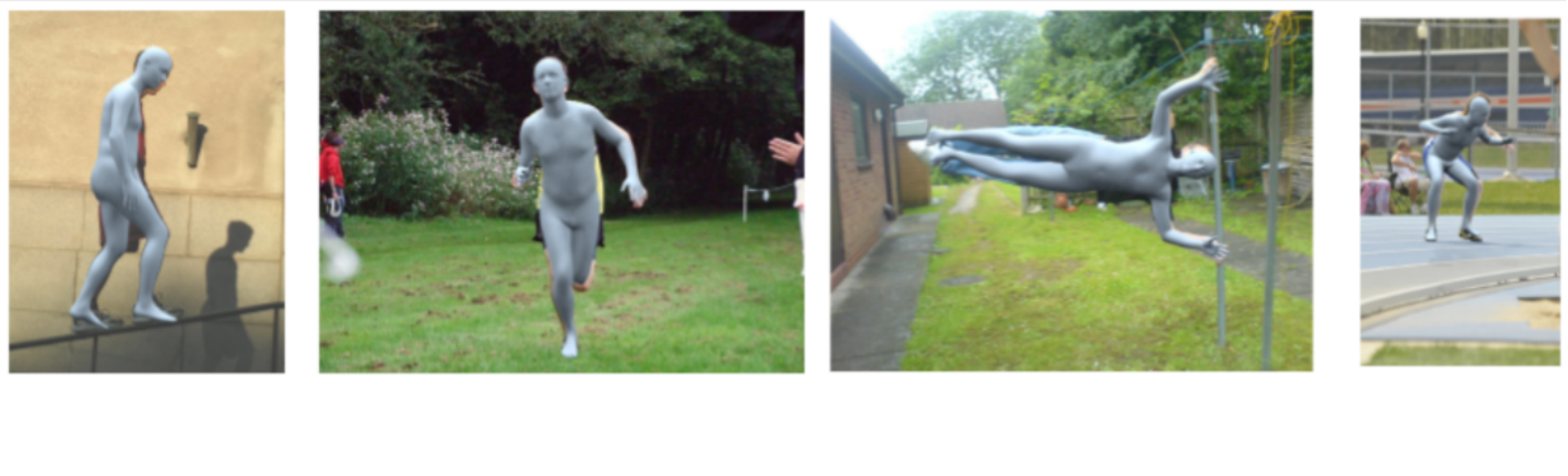}
	\put(7,1){{\parbox{0.5\linewidth}{%
     High resolution LSP dataset~\cite{Johnson10}}}}
\end{overpic}

\begin{overpic}[width=0.48\textwidth,unit=1mm]{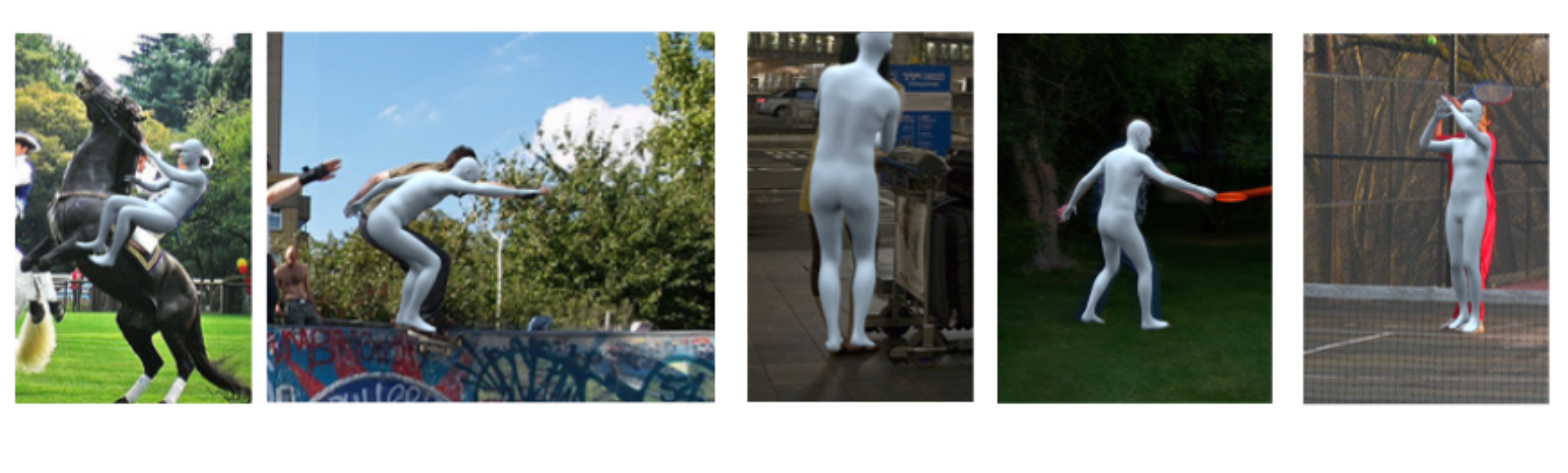}
	\put(15,-1){{\parbox{0.5\linewidth}{%
     COCO dataset~\cite{lin2014microsoft}}}}

\end{overpic}
\begin{overpic}[width=0.48\textwidth,unit=1mm]{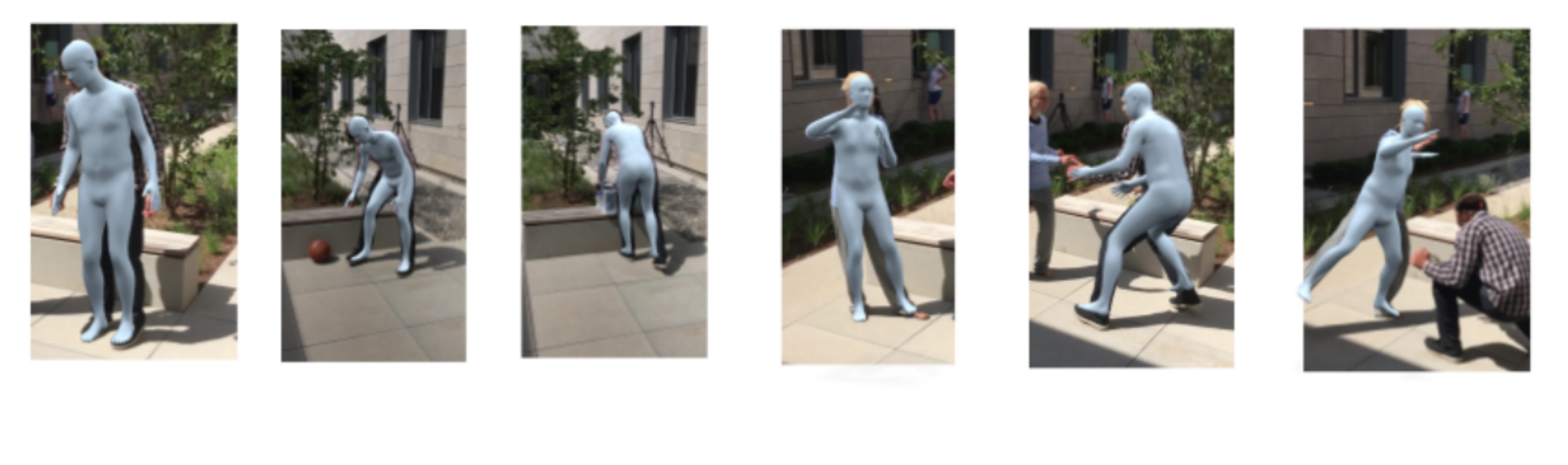}
	\put(15,-1){{\parbox{0.5\linewidth}{%
     3DPW dataset~\cite{vonMarcard2018}}}}
\end{overpic}
\caption{\textbf{3D pose and shape estimation from in-the-wild images} using  \blah{}-based optimization method. }

\label{fig:smplifyx}
\end{figure}

\begin{table}[t]
\centering
\caption{\textbf{3D pose and shape estimation from images} using \blah{}, GAN-S~\cite{Davydov_2022_CVPR} and VPoser~\cite{SMPL-X:2019} as pose prior terms in optimization-based method (\textbf{left}). We also use proposed prior and optimization pipeline to further improve the results of the SoTA 3D pose and shape estimation network, ExPose~\cite{ExPose:2020} (\textbf{right}). } 
\resizebox{\textwidth}{!}{
\begin{tabular}{lcccccccc}
\toprule
Method & \multicolumn{3}{c}{Optimization} & \multicolumn{1}{c}{ExPose} & \multicolumn{4}{c}{ExPose + Optimization}\\
\cmidrule(r){2-4}   \cmidrule(r){6-9}
\multicolumn{1}{c}{} &  VPoser~\cite{SMPL-X:2019}     & GAN-S~\cite{Davydov_2022_CVPR}    & \blah{} &  - & +No prior  &  + VPoser~\cite{SMPL-X:2019}&  + GAN-S~\cite{Davydov_2022_CVPR}    & +\textbf{\blah{}}     \\
\midrule
Per-vertex error ($mm$)  &  60.34 & 59.18 &57.39   &  54.76 & 99.78 & 67.23  &  54.09 & 53.81  \\
\bottomrule
\end{tabular}
\label{tab:imagebased}
}
\end{table}

\subsection{3D pose Estimation from Images}
\label{sec:3dpose}
 We now show that \blah{} can also be used as a prior in optimization-based 3D pose estimation from images~\cite{SMPL-X:2019}. We use the objective function proposed in SMPLify-X~\cite{SMPL-X:2019}, see Eq.~\eqref{eq:3dpose}. Since we are working with a SMPL body only (without hands or faces), we remove the respective loss and prior terms. Thus, we find the desired pose \(\hat{\pose}\) and shape $\hat{\shape}$ as:
\begin{equation}
\label{eq:3dpose}
\hat{\shape}, \hat{\pose}  =  \argmin_{\shape, \pose} \mathcal{L}_J + \lambda_{\pose}\mathcal{L}_{\pose} + \lambda_{\shape}\mathcal{L}_{\shape} + \lambda_{\alpha}\mathcal{L}_{\alpha} \textrm{,}
\end{equation}
with data term $\mathcal{L}_J$, bending term $\mathcal{L}_{\alpha}$, shape regularizer $\mathcal{L}_{\shape}$, and prior term $\mathcal{L}_{\pose}$. The data term and the bending term are given as: 
\begin{equation}
\label{eq:3dpose_indv}
\mathcal{L}_J = \sum_{i \in \mathrm{joints}} \gamma_i w_i \rho(\Pi_K (R_{\theta}(J(\shape))) - J_\mathrm{est, i}) \;  \; \;  \;  \mathcal{L}_{\alpha} = \sum_{i \in (\mathrm{elbow}, \mathrm{knees})} \mathrm{exp}(\pose_i) \textrm{,}
\end{equation} 
where $J_\mathrm{est, i}$ are 2D pose keypoints estimated by a SoTA 2D-pose estimation method~\cite{openpose}, $R_{\theta}$ transforms the joints along the kinematic tree according to the pose $\pose$, $\Pi_K$ represents a 3D to 2D projection with intrinsic camera parameters and $\rho$ represents a robust Geman-McClure error~\cite{Geman1987StatisticalMF}. Further, the bending term $\mathcal{L}_{\alpha}$ penalizes large bending near the elbow and knee joints, and the shape regularizer is given as $\mathcal{L}_{\shape} = || \shape ||^2$~\cite{SMPL-X:2019}. For VPoser, the prior term is given as $\mathcal{L}_{\theta} = ||z||_2^2$, where $z\,$ is the $32$-dimensional latent vector of the VAE. In our model, we use $\mathcal{L}_{\theta} =\udfnet(\pose)$ and minimize the distance of the pose from our learned manifold using our projection algorithm. We leverage the distance information provided by our model in optimization by setting $\lambda_{\pose} = w\udfnet(\pose)$, where $w$ has a fixed value. This ensures that if the pose is getting close to the manifold (\ie $\udfnet(\pose)$ is very small), the prior term is down-weighted, which results in faster convergence.

\noindent\textbf{Results.}
We use the EHF dataset~\cite{SMPL-X:2019} for quantitative evaluation and compare our work with the state-of-the-art priors VPoser~\cite{SMPL-X:2019} and GAN-S~\cite{Davydov_2022_CVPR}. A \blah{} prior term slightly improves on the VPoser and GAN-S based optimization (Tab.~\ref{tab:imagebased}). We observe that the neural network based model ExPose~\cite{ExPose:2020} outperforms all optimization-based results. However, we show that such methods can benefit from an optimization-based refinement step. We refine the ExPose output using Eq~\eqref{eq:3dpose} with \blah{} as prior and compare this refinement with no-prior and other priors (Tab.~\ref{tab:imagebased}). With no prior, the optimization objective only minimizes the joint projection loss, resulting in unrealistic poses. In contrast, GAN and \blah{} improve the result (qualitatively and quantitatively), generating realistic poses, while \blah{} outperforms the GAN prior. 
Finally, in Fig.~\ref{fig:smplifyx} we show qualitative results of optimization-based 3D pose estimation on in-the-wild images from 3DPW~\cite{vonMarcard2018}, LSP~\cite{Johnson10} and MS-COCO~\cite{lin2014microsoft} datasets.

\subsection{Pose Generation} \label{sec:pose_gen}
\begin{figure*}[t]
	\centering
	\begin{overpic}[width=0.42\textwidth,unit=1mm]{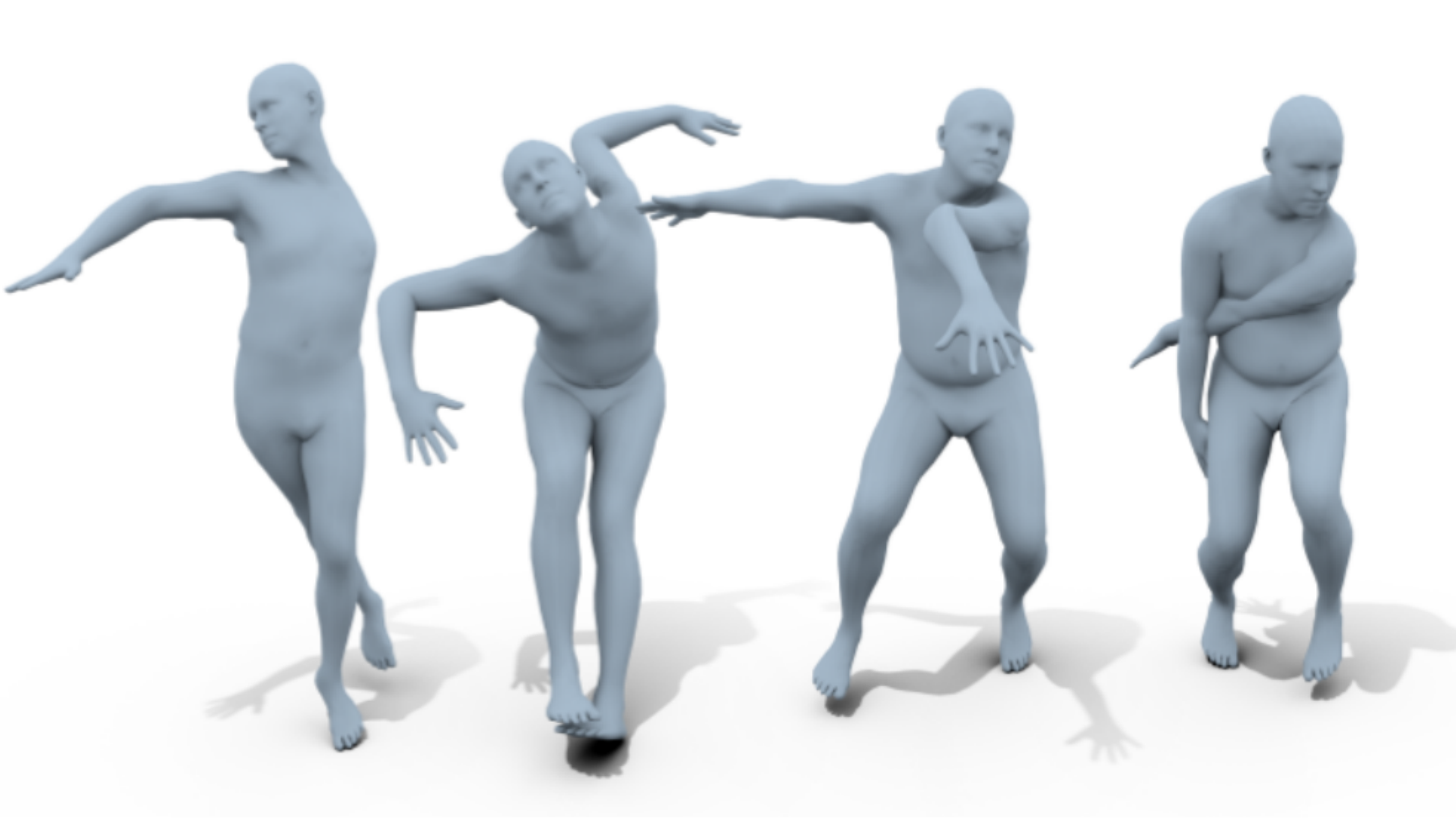}
		\put(24,1){{\parbox{0.4\linewidth}{%
     GMM}}}
    \end{overpic}
    	\begin{overpic}[width=0.42\textwidth,unit=1mm]{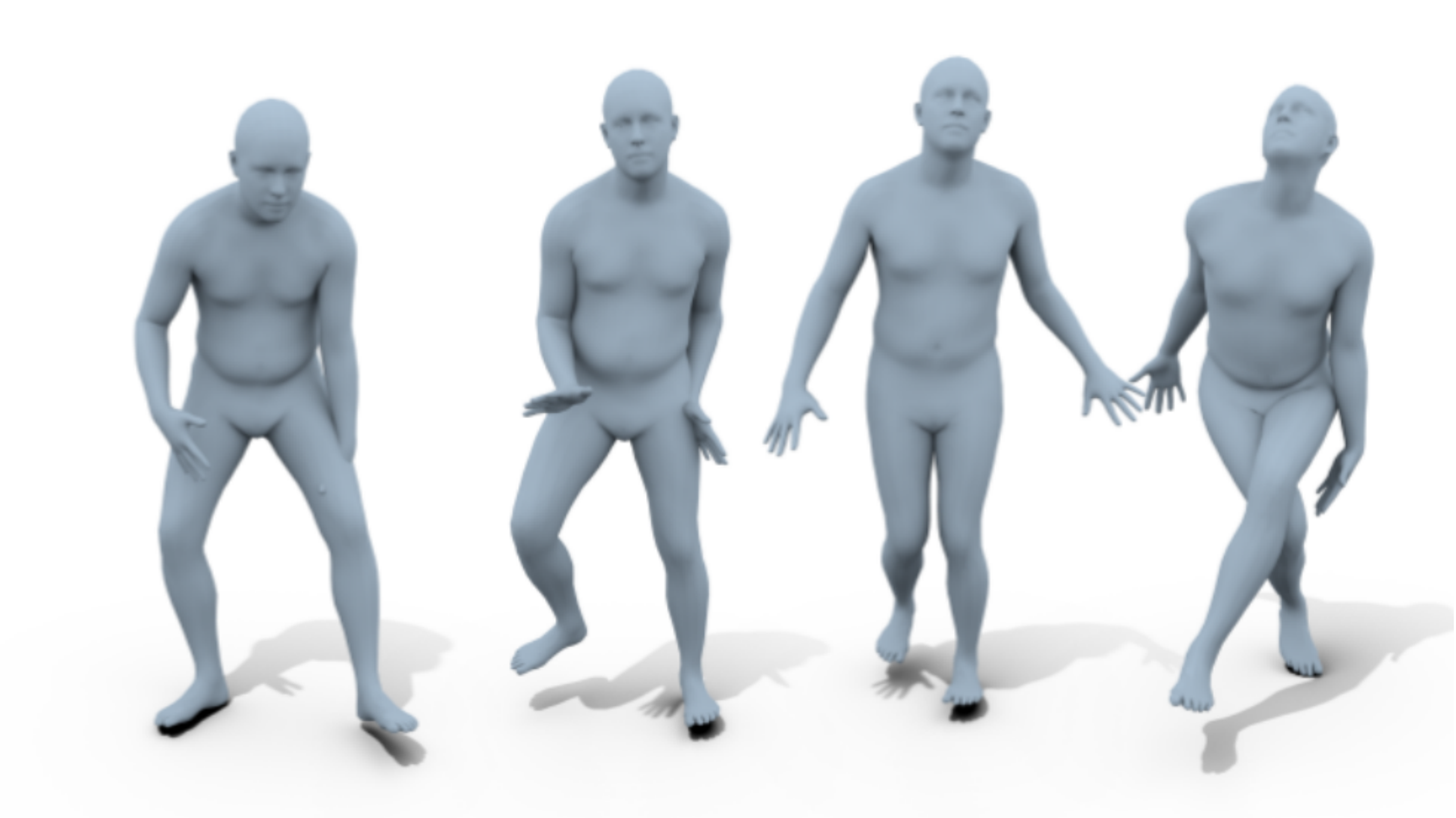}
    			\put(25,1){{\parbox{0.4\linewidth}{%
         VPoser}}}
    \end{overpic}
    
    	\begin{overpic}[width=0.42\textwidth,unit=1mm]{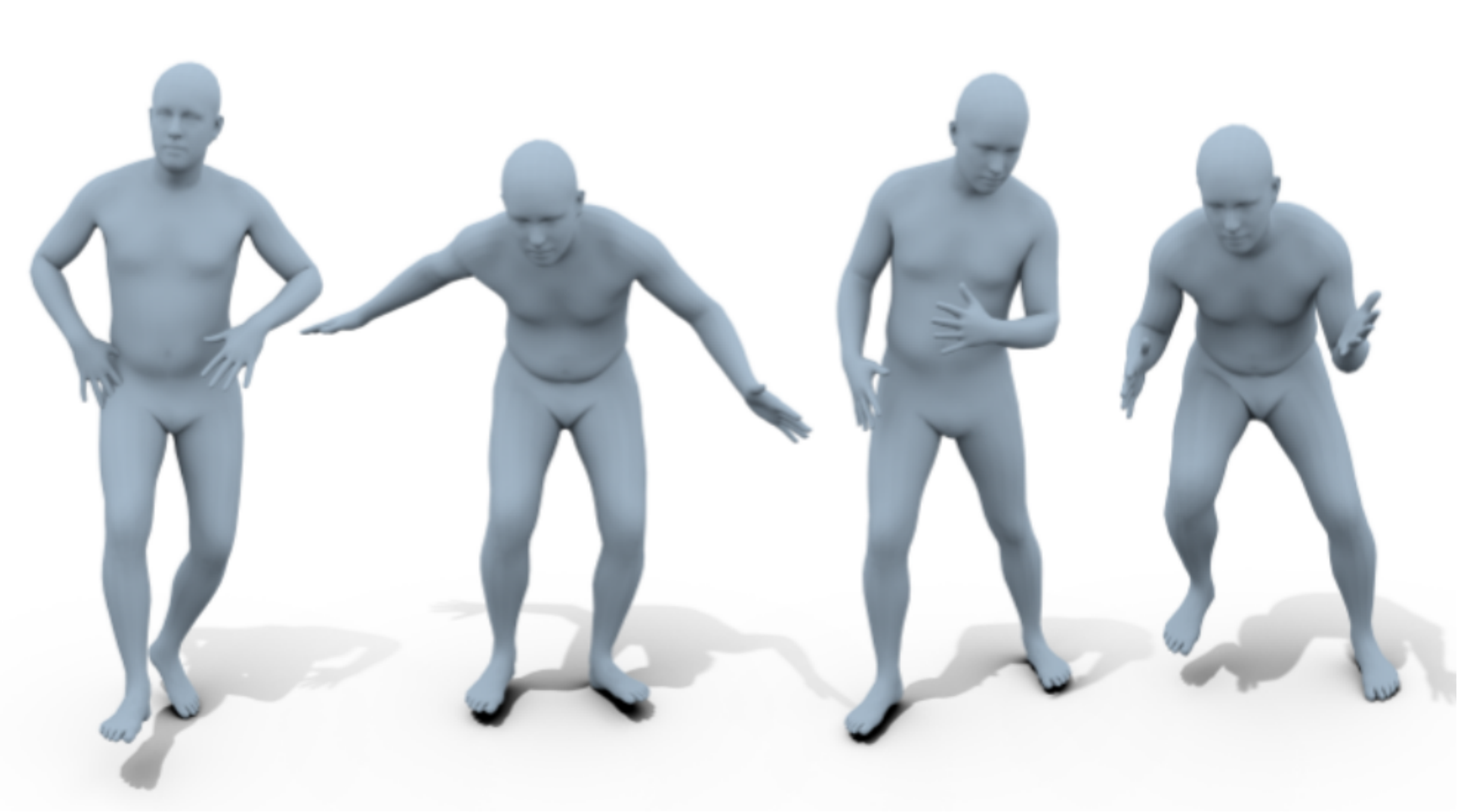}
    			\put(24,0){{\parbox{0.4\linewidth}{%
        GAN-S}}}
    \end{overpic}
    	\begin{overpic}[width=0.42\textwidth,unit=1mm]{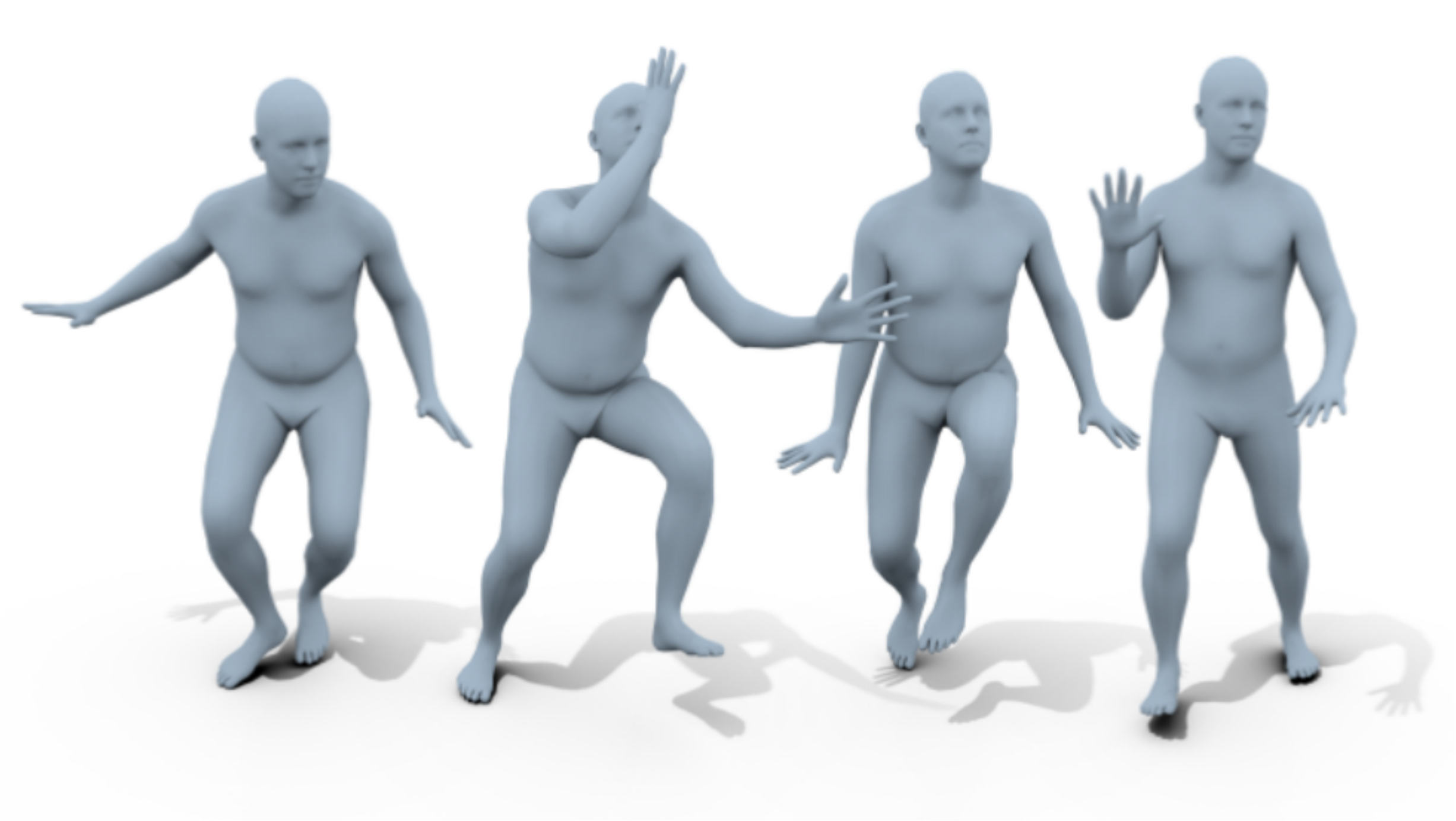}
    			\put(24,0){{\parbox{0.4\linewidth}{%
        Ours}}}
    \end{overpic}
    \caption{\textbf{Pose generation}: GMM generates wrong and unrealistic poses, whereas VPoser, GAN-S and \blah{} generate much more realistic poses. We notice from APD, that variance of poses generated by \blah{} (32.31 cm) is larger than VPoser (23.13 cm) and GAN-S (27.52 cm). } 
	\label{fig:pose_gen}
\end{figure*}

We evaluate our model on the task of pose generation. Due to our distance field formulation, we can generate diverse poses by sampling a random point from $SO(3)^K$ and projecting it onto the manifold (Sec.~\ref{sec:projection}). We compare the results of our model with sampling from the state-of-the-art pose prior VPoser~\cite{SMPL-X:2019}, GMM~\cite{bogo2016keep,omran2018neural} and GAN-S~\cite{Davydov_2022_CVPR} in Fig.~\ref{fig:pose_gen}. We use Average Pairwise Distance (APD)~\cite{stoch_diverse}, to quantify the diversity of generated poses. APD is defined as mean joint distance between all pairs of samples. We randomly sample 500 poses for each GMM, VPoser, GAN-S and \blah{}, which results in APD values of \textbf{48.24}, \textbf{23.13},  \textbf{27.52}, \textbf{32.31} (in cm), respectively. We see that numerically, the GMM produces very large variance, but also results in unrealistic poses, as seen in Fig~\ref{fig:pose_gen} (top-left). \blah{} generates more diverse poses than VPoser while producing only plausible poses. We also calculate the percentage of self-intersecting faces in generated poses, to evaluate one aspect of realism in poses. \blah{} generates poses with less self-intersecting faces ($\textbf{0.89} \%$), as compared to the GAN-S ($\textbf{1.43} \%$) and VPoser  ($\textbf{2.10} \%$).

\subsection{Pose Interpolation}\label{sec:pose_inter}
Pose-NDF learns a manifold of plausible human poses, so it can be used to interpolate between two distinct poses by traversing the manifold. Specifically, for any given pose, we first project start ($\pose_\mathrm{0}$) and end pose ($\pose_\mathrm{T}$) on our manifold using Eq~\eqref{eq:proj}, to get $\pose_\mathrm{0}'$ and  $\pose_\mathrm{T}'$. We then move along the direction of $\pose_\mathrm{T}'$ from $\pose_\mathrm{0}'$ with step size $\tau$ using Eq~\eqref{eq:interpolation}. The interpolated pose ($\pose_\mathrm{t}$) is again projected on the manifold to get a realistic pose ($\pose_\mathrm{t}'$). In the subsequent interpolation steps, we move from $\pose_\mathrm{t}'$ to $\pose_\mathrm{T}'$, where $\pose_\mathrm{t}'$ is updated after each step.

\begin{equation}
\label{eq:interpolation}
\pose_\mathrm{t} = \pose_\mathrm{t-1}' +  \tau (\pose_\mathrm{T}' -\pose_\mathrm{t-1}')
\end{equation}

\begin{figure*}[t]
	\centering
	\begin{overpic}[width=0.8\textwidth,unit=1mm]{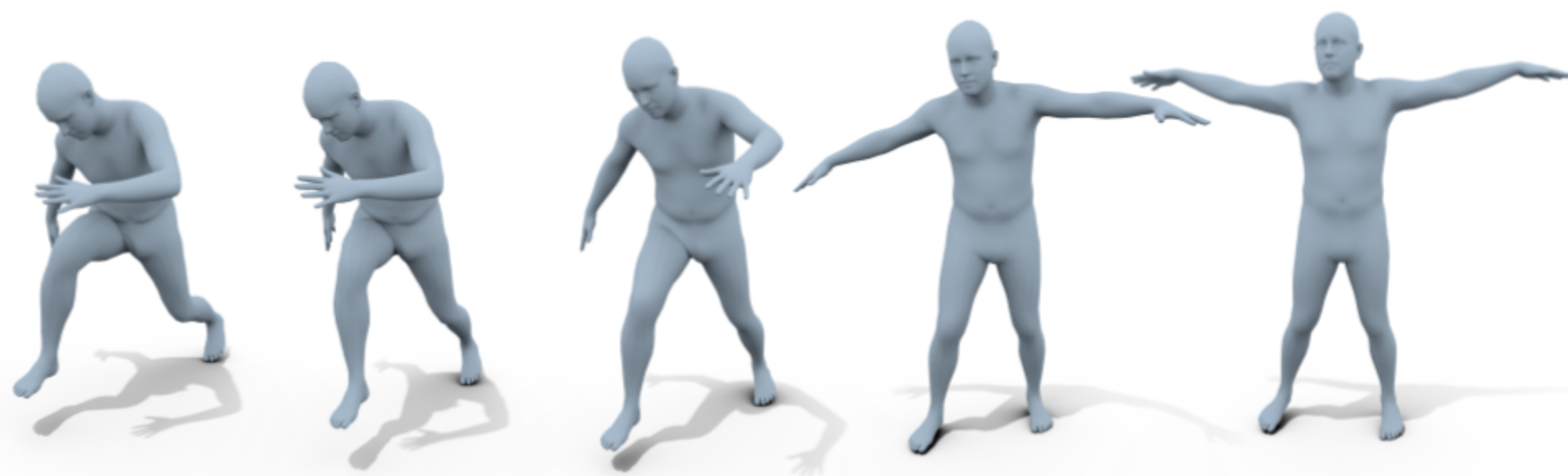}
			\put(-10,15){{\parbox{0.4\linewidth}{% 
      VPoser}}}
    \end{overpic}
	\begin{overpic}[width=0.8\textwidth,unit=1mm]{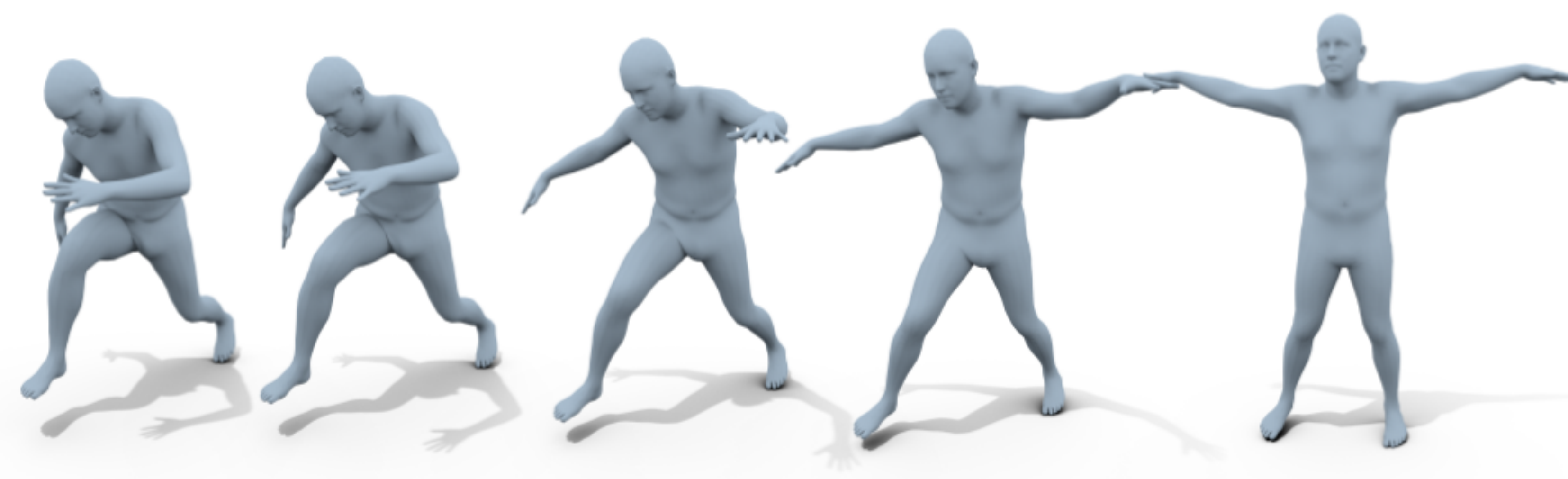}
			\put(-12,15){{\parbox{0.4\linewidth}{%
      \blah{}}}}
    \end{overpic}
	\caption{\textbf{Pose interpolation}: We observe that VPoser-based interpolation (\textbf{top}) is less smooth than \blah{}-based pose interpolation (\textbf{bottom}).}
	\label{fig:interpolation_ex}
\end{figure*}

\noindent\textbf{Results:} We compare the results of \blah{} with those from VPoser~\cite{SMPL-X:2019} and GAN-S~\cite{Davydov_2022_CVPR} interpolation. For VPoser~\cite{SMPL-X:2019}, we project the start and end pose into the latent space and perform linear interpolation using the latent vectors. For GAN-S~\cite{Davydov_2022_CVPR}, we use the spherical interpolation in latent space, as suggested in the work. We qualitatively evaluate the interpolation quality by calculating mean per-vertex distance between consecutive frames. Smaller value means smooth interpolation. We observe that \blah{}-based interpolation has a mean per-vertex distance of 2.72 ± 2.16, GAN-S has 2.71 ± 2.45 and VPoser has 2.53 ± 4.62, which shows that Pose-NDF and GAN-S based interpolation is smooth and the distance in input space is not entirely preserved in case of VAEs. We compare VPoser based interpolation with \blah{} in Fig.~\ref{fig:interpolation_ex}) and observe large jumps in VPoser interpolation. This behaviour is not observed in GAN-S and Pose-NDF based interpolation. Since the VAE learns a compact latent representation of poses, the distance between two input poses is not preserved in the latent space.

\subsection{Pose-NDF vs. Gaussian Assumption models}
\label{sec:vae_gan}
Prior work~\cite{rempe2021humor,SMPL-X:2019} uses VAE-based models as pose/motion prior, which follow a Gaussian assumption in the latent space. This has three major limitations, as mentioned in Sec.~\ref{sec:introduction}. Conversely, \blah{} learns the manifold directly in the pose-space without such assumptions and, hence, overcomes these limitations. 

We report the cumulative error based on deviation from the mean pose. We evaluate on AMASS Noisy (60 and 120 frames) and report cumulative error for samples with $\sigma, 2\sigma, 3\sigma$ for both \blah{} and VPoser motion denoising. We obtain per-vertex error of $\textbf{8.18}, \textbf{8.20}, \textbf{8.21}$ $cm$ for \blah{} and $\textbf{8.35}, \textbf{9.11}, \textbf{9.13}$ $cm$ for VPoser, and  $\textbf{10.08}, \textbf{11.38}, \textbf{16.86}$ $cm$ for HuMoR which reflects that VPoser and HuMoR perform well for poses close to the mean but the error increases for samples deviating from mean pose.
Since the Gaussian distribution is unbounded, it produces \emph{dead regions}, without any data points in these parts of distribution. Hence sampling in these regions might result in completely unrealistic poses for GMM and VPoser (Fig.~\ref{fig:pose_gen}). Lastly, since we learn the manifold in pose space, the distance between individual poses is preserved and leads to smoother interpolation compared to VPoser (see Sec.~\ref{sec:pose_inter}).

\section{Conclusion}
We introduced a novel human pose prior model represented by a scalar neural distance field that describes a manifold of plausible poses as zero level set in $SO(3)^K$. The method extends the idea of classic 3D shape representation using neural fields to higher the dimensions of human poses and maps quaternion-based poses to an unsigned distance value, representing the distance to the pose manifold. The resulting network can be used to project arbitrary poses to the pose manifold, opening applications in several areas. We comprehensively evaluate the performance of our model in diverse pose sampling, pose estimation from images, and motion denoising. We show that our model is able to generate poses with much more diversity than prior VAE-based works and improves state-of-the-art results in reconstruction from images and motion estimation.

\subsubsection{Acknowledgements.} 
Special thanks to the RVH team and reviewers, their feedback helped improve the manuscript and Andrey Davydov, for providing the code for GAN-based pose prior. This work is funded by the Deutsche Forschungsgemeinschaft (DFG, German Research Foundation) - 409792180 (Emmy Noether Programme, project: Real Virtual Humans), German Federal Ministry of Education and Research (BMBF): Tübingen AI Center, FKZ: 01IS18039A  and a Facebook research award. Gerard Pons-Moll is a member of the Machine Learning Cluster of Excellence, EXC number 2064/1 – Project number 390727645.

\clearpage
\bibliographystyle{splncs04}
\bibliography{References}
\end{document}